\documentclass[11pt]{article}

\usepackage[preprint]{acl}
 \usepackage{microtype}

\usepackage[english,bidi=default]{babel} 
\babelfont{rm}{TeXGyreTermesX} 

\usepackage{booktabs} 
\usepackage{multirow} 
\usepackage{graphicx} 
\usepackage{xcolor}
\usepackage{adjustbox}
\usepackage{colortbl}
\usepackage{array}

\usepackage{todonotes}
\usepackage{comment}
\usepackage{amssymb, amsmath}

\title{MIRA--Ev:\\ A Benchmark for Granular Evidence Detection \\and Relational Reasoning in Clinical Exams}

\author{
    Iker De la Iglesia \and Irune Urroz García \and Aitziber Atutxa \and Ander Barrena
    \\ HiTZ Center, University of the Basque Country (EHU)  
    \\ \texttt{\{iker.delaiglesia, aitziber.atucha, ander.barrena\}@ehu.eus}
    \AND 
    Johanna Ramirez-Romero   \and Jose Maria Villa-Gonzalez 
    \\ Hospital Universitario de Cruces 
    \\ \texttt{\{johanna.ramirezromero, josemaria.villagonzalez\}@osakidetza.eus} \\
}

\begin{document}
\maketitle

\begin{abstract}
Clinical NLP evaluation remains dominated by multiple-choice question
answering (MCQA), which scores only final-answer accuracy and cannot detect
when a model reaches the correct diagnosis while grounding it in irrelevant,
absent, or contradictory evidence. 
We introduce \textbf{MIRA-Ev}, a clinical
argument mining benchmark built on Spanish \textit{Médico Interno
Residente} (MIR) licensing-exam cases, re-annotated by expert clinicians
with span-level premises, claims, and directed support/attack relations, and
released in parallel Spanish (native), English, and Basque versions --- the
first clinical argumentation resource in Basque. MIRA-Ev organizes
evaluation into a three-tier task hierarchy: evidence sentence retrieval,
argumentative component extraction, and relation classification.
\end{abstract}

\section{Introduction}

In modern healthcare, the rapid integration of artificial intelligence into diagnostic support has sharpened a long-standing tension: as predictive performance improves, the internal transparency of the systems producing those predictions does not follow. 
In high-stakes clinical settings, this ``black box'' property is not a secondary concern; explainability is a precondition for clinical trust, accountability, and safe deployment. 
Clinical NLP has increasingly looked to Evidence-Based Medicine (EBM) \citep{Goenaga2024Explanatory,YeginbergenA24-corsslingualAM}, the ``conscientious, explicit, and judicious use of current best evidence''\citep{Sackett1996Evidence}, as the normative standard that model reasoning should be held to, and to Clinical Argument Mining (AM) as the formal mechanism for checking whether it is.
By extracting and structuring the premises and claims underlying a clinical decision, AM allows a model to be evaluated not only on \textit{what} it concludes but on \textit{why}.

This distinction between reaching the correct conclusion and reasoning correctly toward it is largely invisible in how clinical language models are currently evaluated.
The dominant paradigm remains multiple-choice question answering (MCQA): benchmarks such as MedQA and MedMCQA score models exclusively on final-answer accuracy. 
Recent work has shown this paradigm is reaching saturation and is only weakly informative about clinical competence: domain-adapted clinical LLMs often fail to outperform general-purpose models on these benchmarks, and comparative evaluations increasingly question whether MCQA performance reflects real diagnostic or predictive capability. 
A model can select the correct option while grounding that selection in irrelevant, absent, or actively contradictory evidence, and current benchmarks cannot detect this failure mode at all because they never inspect the reasoning trace, only its output.

Argument Mining offers a mature formalism for closing this gap, but its application to clinical text remains scarce relative to general-purpose domains like persuasive essays, and its application to \textbf{non-English clinical text is scarcer still}. This scarcity is not incidental. Clinical NLP resource development faces a well-documented double bottleneck: data scarcity compounded by privacy constraints and an evaluation crisis dominated by saturated, English-centric MCQA. This bottleneck is categorically worse for languages that are minority both globally and within the clinical NLP resource landscape specifically. Spanish, despite its size as a world language, remains comparatively a minority in specialized biomedical resources; Basque is severely low-resource by any measure. 

To address this gap, we present \textbf{MIRA-Ev}, a clinical reasoning benchmark built on the Spanish \textit{Médico Interno Residente} (MIR) medical licensing examination, re-annotated with fine-grained, span-level argumentative structure and released as \textbf{parallel Spanish (native), English, and Basque} versions. Each MIR case models differential diagnosis as an argumentative structure: the clinical vignette (patient history, symptoms, examination findings, and laboratory results) supplies scattered medical evidence, \textbf{premises}, that support or attack a set of competing diagnostic hypotheses, \textbf{claims}, typically instantiated by the candidate answer options. Critically, this mapping is not fixed: in a subset of cases, an answer option itself functions as a premise, and the claim it bears on is instead embedded within the case or question stem alongside other evidence. This inversion is deliberately preserved rather than normalized away, because it lets us test whether a system infers argumentative role from content and context, rather than from the positional heuristic ``options are claims'', a heuristic we expect many models, particularly non-fine-tuned LLMs, to implicitly (and incorrectly) rely on.

Constructing and evaluating this resource surfaces three computational bottlenecks that we take as defining clinical argument mining as a task family, distinct from its persuasive-text origins:

\begin{enumerate}
    \item \textbf{Fine-grained, sub-sentential span extraction.} Roughly 20\% of evidence spans in MIRA-Ev are sub-sentential clauses delimited by punctuation or contrastive conjunctions rather than sentence boundaries, which standard sentence-level splitters cannot recover, creating a hard performance ceiling under strict boundary-matching evaluation.
    \item \textbf{Cascading pipeline error.} The task is structured as a hierarchical pipeline: sentence relevance, span extraction, and directed relation classification, in which relation detection is conditioned on the correctness of upstream boundaries.
    \item \textbf{Argumentative role ambiguity.} Because the claim/premise mapping is not fixed to a syntactic position (case vs. option), models cannot rely on positional shortcuts and must instead infer argumentative function from content, a capability we can directly probe via the inverted-mapping subset.
\end{enumerate}

We evaluate this benchmark using both a domain-specific Spanish/Basque bilingual encoder, EriBERTa \citep{eriberta}, and a range of medical-pretrained and general-purpose generative LLMs, establishing empirical bounds across all three tiers of the task and across languages.

\textbf{Contributions:} 
\begin{itemize}
    \item [(i)] MIRA-Ev, a clinical argument mining resource built on MIR exam cases with expert clinician re-annotation at word-level span granularity, released in parallel Spanish/English/Basque versions; 
    \item [(ii)] a reframing of MCQA-sourced exam data as a structured evidence-and-relation task that decouples answer correctness from reasoning validity; 
    \item [(iii)] a three-tier evaluation hierarchy (sentence retrieval, component extraction, relation classification) with matched strict/relaxed metrics; 
    \item [(iv)] empirical benchmarking of an encoder pipeline and LLMs across tiers, languages, and the standard/inverted argumentative mapping; 
    \item [(v)] a contribution to low-resource clinical NLP, with Basque explicitly foregrounded rather than treated as a translation afterthought.
\end{itemize}

\section{Related Work} 

\subsection{Argument Mining in Persuasive and Medical Domains}

Argument Mining was initially developed for well-structured, user-generated text such as persuasive essays, where explicit discourse markers make claim/premise identification comparatively tractable; classic systems modeled component identification and directed relations via sequential tagging (BiLSTM/CRF architectures) \citep{Yeginbergen2024Argument,GarciaFerrero2022Model}. Its extension to biomedical text represents a substantial increase in discursive complexity. Early medical AM work targeted randomized controlled trial abstracts (e.g., the AbstRCT corpus \citep{Mayer2021Enhancing}, MEDLINE-derived)  \citep{WHO2021ClinicalTrials}, combining domain-pretrained transformers with LSTM/GRU/CRF layers \citep{campillos-llanos-etal-2021-clinical,ZakhirPuig2026ClinArgES}, and established two findings that MIRA-Ev inherits as design constraints: domain-specific pretraining is necessary to capture clinical terminology, and span boundary detection, not relation classification, is the primary bottleneck, since misaligned spans propagate errors downstream.

\subsection{Evidence Extraction in Biomedical Literature: PICO and EBM Sentence Classification} 

A closely related but distinct line of work extracts evidentiary structure from biomedical \textit{literature} rather than clinical \textit{case narratives}. PICO extraction identifies Population, Intervention, Comparison, and Outcome elements within RCT abstracts, and EBM sentence classification assigns categorical discourse roles (e.g., background, methods, results) to sentences within structured abstracts. Both operate over single, self-contained documents and produce categorical or extractive labels without modeling directed argumentative relations between components. MIRA-Ev differs along three axes: (i) it operates over clinical case vignettes paired with competing diagnostic hypotheses rather than literature abstracts; (ii) it requires directed, polarity-labeled relations (support/attack) between premises and claims rather than sentence-level categorical labels; (iii) its evidence spans are annotated at sub-sentential, token-level granularity rather than sentence level. MIRA-Ev is consequently closer in spirit to relational argument mining than to PICO-style information extraction, despite superficial similarity in the ``find the evidence'' framing.

\subsection{Clinical NLP and Argumentative Corpora in Spanish and Basque}

Clinical NLP in non-English settings suffers from persistent resource scarcity, motivating cross-lingual transfer as an active research direction. In Spanish, several argumentation-driven resources have recently emerged: the Antidote CasiMedicos corpus \citep{antidote} introduced MIR exam questions enriched with gold explanatory arguments authored by volunteer clinicians; this was extended to CasiMedicos-Arg \citep{casimedicos-arg}, a multilingual medical QA dataset with gold explanatory argument structures across Spanish, English, French, and Italian; and ClinArgES provides sentence-level Spanish clinical argumentation with explicit subject-predicate structure.

MIRA-Ev builds on this lineage but departs from it on two axes simultaneously: it shifts annotation granularity from sentence-level argument structure to token-level span extraction with directed relation typing, and it extends the language coverage to include \textbf{Basque}, a language with essentially no prior clinical argumentation resources at any granularity. This addresses a resource gap that is qualitatively different from the Spanish case; Basque clinical NLP faces not comparative under-resourcing but near-total absence of annotated reasoning data, making its inclusion here a primary contribution.

\subsection{The Evaluation Crisis: Beyond Multiple-Choice QA in Clinical NLP}

A growing body of work questions whether MCQA-style benchmarks meaningfully measure clinical competence. Domain-adapted clinical LLMs frequently fail to outperform general-purpose models on standard medical MCQA \citep{jeong2024,dorfner2024}, and comparative studies find traditional ML models remain competitive with LLMs on clinical prediction tasks despite MCQA gains \citep{chen2025}, jointly suggesting that MCQA accuracy has become decoupled from the clinical capabilities it is assumed to proxy \citep{delaiglesia-ranking-over-scoring}. This evaluation crisis is amplified for minority languages, where native, expert-curated benchmarks are exceptionally rare, and evaluation typically proceeds via translated English resources, introducing confounds between clinical and translation performance \citep{domingo-aldama2026}. MIRA-Ev is positioned directly against this gap: rather than another MCQA resource, it evaluates the \textit{evidentiary grounding} of a model's answer, independently of whether the answer itself is correct, and does so natively in a minority language rather than through translation.

\section{The MIRA-Ev Dataset}

In this section, we introduce MIRA-Ev, a benchmark for fine-grained evidence detection and relational reasoning over clinical case narratives. We first describe the three-tier task hierarchy that structures the dataset, then the source data and de-annotation procedure used to construct it, its parallel multilingual release across Spanish, English, and Basque, and finally the resulting corpus statistics.

\subsection{Task Hierarchy}

MIRA-Ev structures clinical reasoning as a hierarchical, cascading task sequence across three tiers.

\subsubsection{Task 1: Evidence Sentence Retrieval}
A binary sentence-level classification task isolating decision-critical information from narrative context: given a case, identify which sentences contain findings relevant to evaluating the diagnostic options, as opposed to clinical background noise.

\subsubsection{Task 2: Argumentative Component Extraction}
A token-level extraction task recovering exact boundaries of argumentative
components within relevant sentences, typed as:
\begin{itemize}
    \item \textbf{Premises}, granular clinical observations: symptoms, exam
    findings, lab/imaging values.
    \item \textbf{Claims}, explicit diagnostic conclusions or hypotheses.
\end{itemize}

By default, premises are instantiated in the case, and claims by the
candidate options, but this mapping inverts in a subset of instances, where an
option functions as a premise, and the corresponding claim is embedded in the
case/question stem alongside other supporting or attacking evidence. Systems
are expected to identify component \textit{type} from content and
argumentative function, not from positional location (case vs. option).

\subsubsection{Task 3: Relation Classification}
Directed relation classification between components and candidate options:
\begin{itemize}
    \item \textbf{Support}, evidence indicates the option is correct/appropriate.
    \item \textbf{Attack}, evidence contradicts, rules out, or invalidates the option.
\end{itemize}

\subsection{Source Data and De-Annotation}

Built from official Spanish MIR exams sourced from CasiMedicos; prior
annotations and clinician explanations were stripped to retrieve raw cases
(patient history, presenting symptoms, clinical query, and a mix of 4--5
option cardinalities across items). Splits are stratified by medical
specialty and question type.

\subsection{Multilingual Construction}

MIRA-Ev is a Spanish native dataset with parallel English and Basque
versions.

\subsection{Corpus Statistics}

\begin{table}[ht]
\centering
\adjustbox{max width=\columnwidth}{
\begin{tabular}{lcccc}
\toprule
\textbf{Corpus Element} & \textbf{Train} & \textbf{Dev} & \textbf{Test} & \textbf{Total} \\
\midrule
Cases & 128 & 24 & 102 & \textbf{254} \\
Not-Relevant Sentences & 263 & 41 & 171 & \textbf{475} \\
Relevant Sentences & 237 & 61 & 222 & \textbf{520} \\
\midrule
Premises & 572 & 121 & 475 & \textbf{1,168} \\
Claims & 586 & 110 & 479 & \textbf{1,175} \\
\midrule
Support Relations & 613 & 133 & 661 & \textbf{1,407} \\
Attack Relations & 646 & 138 & 571 & \textbf{1,355} \\
\bottomrule
\end{tabular}}
\caption{Detailed corpus statistics for the MIRA-Ev dataset.}
\label{tab:dataset_stats}
\end{table}

\section{Experimental Setup}
\subsection{Automatic Evaluation}

Systems are assessed using macro-averaged Precision ($P$), Recall ($R$), and
$F_1$-score as the fundamental metrics across all three tasks. For
extraction tasks, matching follows a strictly one-to-one mapping rule, where
a unique gold-standard target can be linked to at most one predicted
counterpart. System performance is categorized and evaluated under three
distinct algorithmic task types:

\paragraph{Sentence-Level Classification}
It is formulated as a binary classification problem tasked with
distinguishing critical clinical evidence from narrative background context.
Performance is reported across both categories (\emph{relevant} and
\emph{not-relevant}).

\paragraph{Argument Component Detection}
Systems must identify textual spans and categorize them according to their
argumentative function. Evaluation utilizes two specific operational
matching criteria:
\begin{description}
    \item \textbf{Strict Matching:} Requires an exact, character-level
    alignment of boundary offsets between the predicted span and the
    gold-standard reference.
    \item \textbf{Relaxed Matching:} Evaluates token-level overlap using the
    Jaccard Index ($J$), accommodating minor boundary variations (e.g.,
    leading articles or trailing punctuation) without penalizing them as
    full mismatches. A predicted span is considered a valid match if it
    meets a fixed intersection-over-union (IoU) threshold $\tau$
    (where $\tau = 0.5$):
    $$J(A, B) = \frac{|A \cap B|}{|A \cup B|} \geq \tau$$
\end{description}

\paragraph{Argumentative Relation Detection}
Systems are required to predict directed links between components. A
predicted relation is registered as a true positive if and only if there is
simultaneous agreement on the relation type (polarity) and the exact
structural identities (source and target spans) of both arguments. This
layer is reported using both strict and relaxed matching criteria mapped
from the underlying components.

\subsection{Encoder Methodology}
\label{sec:encoder_methodology}
We establish encoder benchmarks by approaching the hierarchical pipeline
sequentially using a pre-trained bilingual EriBERTa-base encoder model \citep{eriberta}.

\subsubsection{Task 1: Evidence Sentence Retrieval}
Formulated as context-aware token classification. Case sentences are
concatenated with candidate options into a unified input string:
\begin{equation*}
\tiny
\texttt{[CLS] } S_0 \texttt{ [SEP] } S_1 \texttt{ [SEP] } \dots \texttt{ [SEP] } S_N \texttt{ [SEP] options [SEP]}
\end{equation*}
A linear classification head is applied to the hidden representation of each
sentence-demarcating \texttt{[SEP]} token to predict its binary relevance.

\subsubsection{Task 2: Argumentative Component Extraction}
Formatted as sequence tagging using the BIOES scheme over \texttt{Premise},
\texttt{Claim}, and \texttt{O} classes. To handle subword tokenization, only
the first subword of each term receives its gold-standard label, while
subsequent subwords are masked during loss calculation.

\subsubsection{Task 3: Relation Classification}
Employs entity-marker encoding. Specific markers (\texttt{[E1]/[/E1]} and
\texttt{[E2]/[/E2]}) are injected directly into the case text to bracket the
predicted spans. The pooled \texttt{[CLS]} representation is fed to a
classification head to predict \texttt{Support}, \texttt{Attack}, or
\texttt{None}.

\section{Results}
We evaluate the sequential encoder pipeline described in
Section~\ref{sec:encoder_methodology} across the three MIRA-Ev tasks, reporting
both strict and relaxed span-matching criteria as defined above. Overall
pipeline performance is 47.90, a score that is shared across all three
tables because it reflects the end-to-end pipeline run from which each
task-level breakdown is derived, rather than an independent per-task metric.
Table~\ref{tab:task1_results} reports results for Task 1 (Evidence Sentence
Retrieval); Table~\ref{tab:task2_results} for Task 2 (Argumentative
Component Extraction); and Table~\ref{tab:task3_results} for Task 3
(Relation Classification). Because Task 1 is a sentence-level classification
task with no span boundaries to match, its strict and relaxed scores are
identical throughout, which serves as an internal consistency check on the
evaluation protocol rather than a substantive finding.

For Task 1 (Table~\ref{tab:task1_results}), the pipeline reaches 74.81\%
accuracy and a macro $F_1$ of 73.48 ($P=75.31$, $R=73.13$). Performance is
higher on the Relevant class ($F_1=79.42$) than on the Not-Relevant class
($F_1=67.54$), a gap driven mainly by recall (86.04 vs.\ 60.23) rather than
precision (73.75 vs.\ 76.87), indicating a tendency to over-predict sentence
relevance rather than to under-predict it.

Task 2 (Table~\ref{tab:task2_results}) shows a substantial gap between
strict and relaxed matching, with macro $F_1$ rising from 62.63 to 68.02.
This gain is driven almost entirely by Premise extraction ($F_1$: 27.03
$\rightarrow$ 37.80), whereas Claim extraction is essentially unaffected by
boundary strictness ($F_1=98.23$ under both criteria) --- consistent with
claims being recoverable via a near-exact match to the candidate options
rather than requiring free-text span identification.

Task 3 (Table~\ref{tab:task3_results}) yields the lowest performance in the
pipeline, with macro $F_1$ of 7.58 under strict matching and 10.61 under
relaxed matching. In this evaluation, \textit{Attack} relations are detected
with higher $F_1$ than \textit{Support} relations under both criteria
(strict: 8.84 vs.\ 6.32; relaxed: 12.98 vs.\ 8.24), a pattern driven largely
by higher recall on Attack (16.81 vs.\ 6.51 strict).

\begin{table}[htbp]
\centering
\setlength{\tabcolsep}{3pt}
\resizebox{\linewidth}{!}{%
\begin{tabular}{l l c *{10}{c}}
\toprule
\multirow{3}{*}{\textbf{Boundary}} & \multirow{3}{*}{\textbf{Pipeline}} & \multirow{3}{*}{\textbf{Overall}} &
\multicolumn{10}{c}{\textbf{Sentence Relevance}} \\
\cmidrule(lr){4-13}
& & &  & \multicolumn{3}{c}{Macro} & \multicolumn{3}{c}{Relevant} & \multicolumn{3}{c}{Not Rel.} \\
\cmidrule(lr){5-7} \cmidrule(lr){8-10} \cmidrule(lr){11-13}
& & & Acc & F1 & P & R & F1 & P & R & F1 & P & R \\
\midrule
Strict & Encoder + Classifier & 47.90 & 74.81 & 73.48 & 75.31 & 73.13 & 79.42 & 73.75 & 86.04 & 67.54 & 76.87 & 60.23 \\
Relaxed & Encoder + Classifier & 47.90 & 74.81 & 73.48 & 75.31 & 73.13 & 79.42 & 73.75 & 86.04 & 67.54 & 76.87 & 60.23 \\
\bottomrule
\end{tabular}%
}
\caption{Detailed results for Task 1 (Evidence Sentence Retrieval).}
\label{tab:task1_results}
\end{table}

\begin{table}[htbp]
\centering

\setlength{\tabcolsep}{3pt}
\resizebox{\linewidth}{!}{%
\begin{tabular}{l l c *{9}{c}}
\toprule
\multirow{3}{*}{\textbf{Boundary}} & \multirow{3}{*}{\textbf{Pipeline}} & \multirow{3}{*}{\textbf{Overall}} &
\multicolumn{9}{c}{\textbf{Component Detection}} \\
\cmidrule(lr){4-12}
& & & \multicolumn{3}{c}{Macro} & \multicolumn{3}{c}{Premise} & \multicolumn{3}{c}{Claim} \\
\cmidrule(lr){4-6} \cmidrule(lr){7-9} \cmidrule(lr){10-12}
& & & F1 & P & R & F1 & P & R & F1 & P & R \\
\midrule
Strict & Encoder + Classifier & 47.90 & 62.63 & 62.03 & 63.27 & 27.03 & 26.13 & 28.00 & 98.23 & 97.93 & 98.54 \\
Relaxed & Encoder + Classifier & 47.90 & 68.02 & 67.23 & 68.85 & 37.80 & 36.54 & 39.16 & 98.23 & 97.93 & 98.54 \\
\bottomrule
\end{tabular}%
}
\caption{Detailed results for Task 2 (Argumentative Component Extraction), strict and relaxed matching.}
\label{tab:task2_results}
\end{table}

\begin{table}[htbp]
\centering

\setlength{\tabcolsep}{3pt}
\resizebox{\linewidth}{!}{%
\begin{tabular}{l l c *{9}{c}}
\toprule
\multirow{3}{*}{\textbf{Boundary}} & \multirow{3}{*}{\textbf{Pipeline}} & \multirow{3}{*}{\textbf{Overall}} &
\multicolumn{9}{c}{\textbf{Relation Detection}} \\
\cmidrule(lr){4-12}
& & & \multicolumn{3}{c}{Macro} & \multicolumn{3}{c}{Support} & \multicolumn{3}{c}{Attack} \\
\cmidrule(lr){4-6} \cmidrule(lr){7-9} \cmidrule(lr){10-12}
& & & F1 & P & R & F1 & P & R & F1 & P & R \\
\midrule
Strict & Encoder + Classifier & 47.90 & 7.58 & 6.07 & 11.66 & 6.32 & 6.15 & 6.51 & 8.84 & 5.99 & 16.81 \\
Relaxed & Encoder + Classifier & 47.90 & 10.61 & 8.41 & 16.58 & 8.24 & 8.01 & 8.47 & 12.98 & 8.80 & 24.69 \\
\bottomrule
\end{tabular}%
}
\caption{Detailed results for Task 3 (Relation Classification).}
\label{tab:task3_results}
\end{table}

\section{Discussion}

We now analyze the experimental results in light of the structural
properties of MIRA-Ev. Four findings stand out: the extent to which claim
detection performance reflects structural shortcuts rather than genuine
extraction ability; the gap between strict and relaxed boundary matching;
the propagation of upstream extraction errors through the relation
classification pipeline; and a persistent asymmetry between support and
attack relation detection.

\paragraph{Claim Detection: Reliance on Structural Heuristics}
The high performance observed in claim detection is largely an artifact of
the dataset's structural design rather than a reflection of genuine
extraction capabilities among the participating systems. Because the
candidate options typically map directly to the claims within the defined
task ontology, models effectively reduced this subtask to a simple lookup
operation. System performance degraded sharply, however, in the minority of
cases where the functional roles reversed --- specifically, when the claim
was embedded within the clinical narrative and the candidate options served
as evidence. Current architectures struggled in these instances because they
relied on positional heuristics (defaulting to options as claims) rather
than inferring the true functional roles of the text. In contrast, premise
extraction required identifying fragmented clinical observations within
free-text narratives, making it a far more reliable indicator of true system
capability in this task.

\paragraph{Boundary Sensitivity vs. Semantic Comprehension}
The substantial performance gains observed when shifting from strict to
relaxed evaluation metrics indicate that models successfully captured the
relevant clinical semantics but were heavily penalized for boundary noise.
Systems frequently lost credit for superficial token inclusion --- such as
trailing punctuation, conjunctions, or leading articles --- rather than for
failing to identify the underlying clinical finding itself.

\paragraph{Error Propagation in Relation Classification}
The degraded performance in Task 3 is better characterized as a cascading
pipeline error rather than an intrinsic failure of the relation
classification models. Because the evaluation strictly mandated exact
boundary matching for the source span, any minor token misalignment
generated during the premise extraction phase propagated downstream,
invalidating what would otherwise be correctly typed relations.

\section{Ethics Statement}
It is imperative to note that these models are intended strictly for research and evaluation purposes. The persistent vulnerabilities observed in handling negation and complex deduction underscore the risk of deploying such architectures in real-world clinical decision-making environments without extensive human oversight.

\section{Conclusion}
The MIRA-Ev dataset establishes a rigorous benchmark for evaluating Spanish clinical argumentation, exposing critical areas for improvement in future NLP systems. Our analysis reveals that rigid boundary-matching criteria often obscured the models' actual semantic comprehension, turning minor span misalignments into significant evaluation penalties. 
Furthermore, the persistent performance gap between detecting \textit{Attack} versus \textit{Support} relations highlights a universal vulnerability in processing negation and comparative logic within clinical contexts. 
Overcoming these bottlenecks in boundary flexibility and complex reasoning will be essential for advancing automated clinical extraction pipelines in subsequent shared tasks.

\section{Acknowledgments}
This work is supported by MCIN/AEI/ 10.13039/501100011033 and FEDER/UE via EDHER-MED -- EDHIA (PID2022-136522OB-C22), TRUST-MED (PID2025-174880OB-I00) and DeepKnowledge (PID2021-127777OB-C21). 
Secondary support is provided by the HiTZ Center and the Basque Government (IT1570-22), and EFA 104/01-LINGUATEC IA (ERDF/INTERREG POCTEFA 2021-2027). 
I. de la Iglesia is funded by an MCIU FPU grant (FPU23/03347).


\bibliography{bibliography}


\end{document}